\def\year{2022}\relax
\documentclass[letterpaper]{article} 
\usepackage{aaai22}  
\usepackage{times}  
\usepackage{helvet}  
\usepackage{courier}  
\usepackage[hyphens]{url}  
\usepackage{graphicx} 
\usepackage{natbib}  
\usepackage{caption} 
\DeclareCaptionStyle{ruled}{labelfont=normalfont,labelsep=colon,strut=off} 
\usepackage{algorithm}
\usepackage{algorithmic}

\usepackage{newfloat}
\usepackage{listings}
\floatstyle{ruled}
\newfloat{listing}{tb}{lst}{}
\floatname{listing}{Listing}
\usepackage{amsmath}
\usepackage[inline]{enumitem}

\title{Zero-Shot and Few-Shot Classification of Biomedical Articles \\ in Context of the COVID-19 Pandemic}
\author {
    Simon Lupart\textsuperscript{\rm 1}\footnote{Work conducted during an internship at AMU, CNRS, LIS / Marseille in 2021.}
    Benoit Favre\textsuperscript{\rm 2}
    Vassilina Nikoulina\textsuperscript{\rm 3}
    Salah Ait-Mokhtar\textsuperscript{\rm 3}
}
\affiliations {
    \textsuperscript{\rm 1} Grenoble INP - Ensimag\\
    \textsuperscript{\rm 2} Aix Marseille Univ, CNRS, LIS / Marseille\\
    \textsuperscript{\rm 3} NAVER LABS Europe\\
    simon.lupart@gmail.com, benoit.favre@lis-lab.fr, vassilina.nikoulina@naverlabs.com, \\ salah.ait-mokhtar@naverlabs.com
}

\begin{document}

\maketitle

\begin{abstract}

MeSH (Medical Subject Headings) is a large thesaurus created by the National Library of Medicine and used for fine-grained indexing of publications in the biomedical domain. In the context of the COVID-19 pandemic, MeSH descriptors have emerged in relation to articles published on the corresponding topic. Zero-shot classification is an adequate response for timely labeling of the stream of papers with MeSH categories.
In this work, we hypothesise that rich semantic information available in MeSH has potential to improve BioBERT representations and make them more suitable for zero-shot/few-shot tasks. We frame the problem as determining if MeSH term definitions, concatenated with paper abstracts are valid instances or not, and leverage multi-task learning to induce the MeSH hierarchy in the representations thanks to a seq2seq task.
Results establish a baseline on the MedLine and LitCovid datasets, and probing shows that the resulting representations convey the hierarchical relations present in MeSH.

\end{abstract}

\section{Introduction}

\begin{figure}
	\includegraphics[width=0.9\columnwidth]{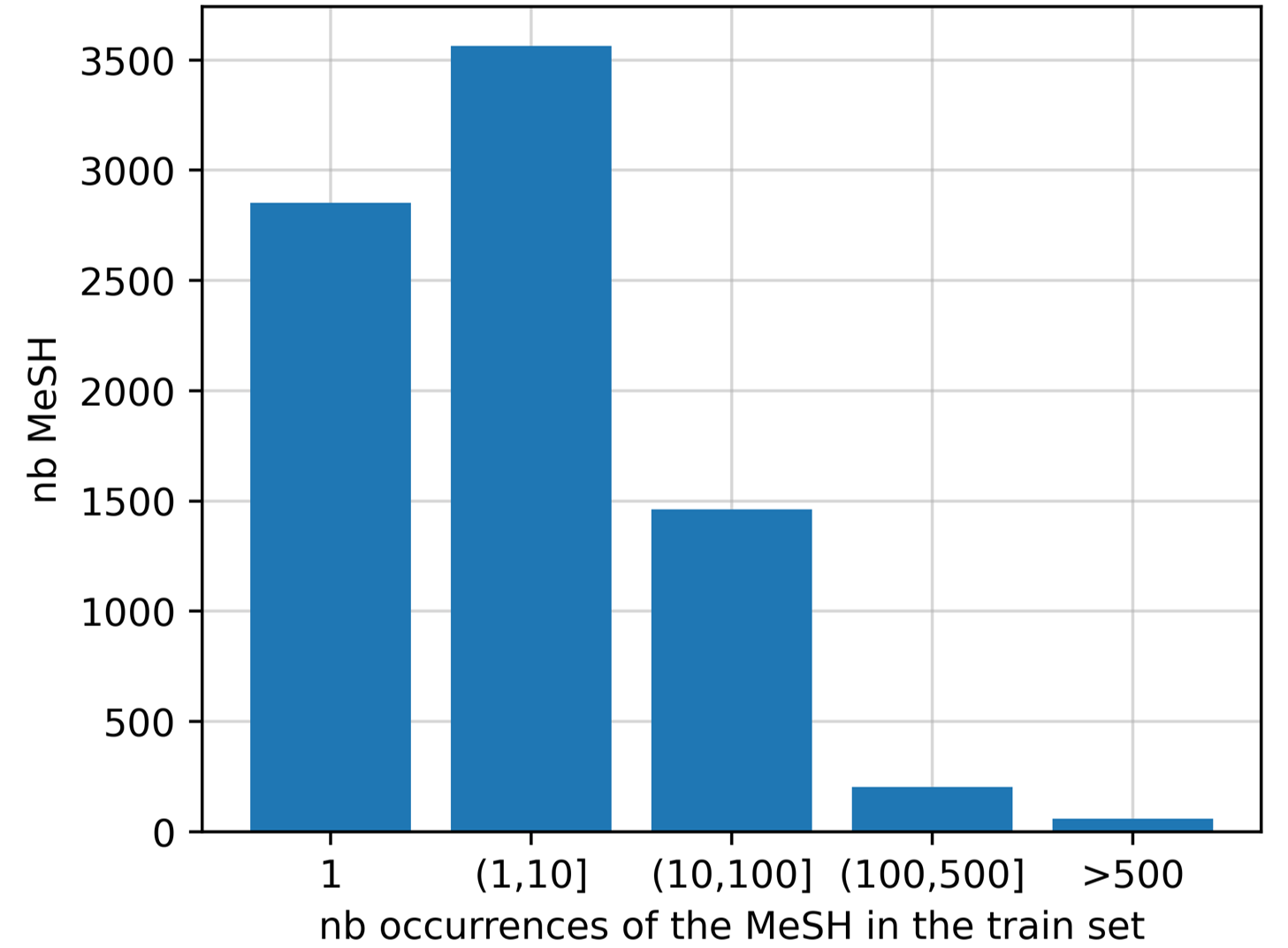}
	\caption{Distribution of the MeSH descriptors: 2853 MeSH terms appear only once in the train dataset, 3565 between 2 and 10 times, while a minority of them appear very frequently (out of 19,125 annotated documents with 8,140 MeSH descriptors). }
	\label{fig:mesh_dist}
\end{figure}

With the outbreak of the COVID-19 disease, the biomedical domain has evolved: new concepts have emerged, and old ones have been revised. In that context, scientific papers are typically manually or automatically labelled with MeSH terms, Medical Subject Headings~\cite{fb1963medical}, which helps routing them to the best target audience. It is crucial for the community to be able to react swiftly to events like pandemics, and manual efforts to annotate large numbers of publications may not be timely. To automate that task, it is difficult to use typical classification methods because of the lack of data for some classes, we therefore consider this problem as a zero-shot/few-shot documents classification problem. Formally, in zero-shot learning, at test time a learner (the model) observes documents of classes that were not seen during training, and respectively in few-shot learning, the model will have seen only a small number of documents with these classes. Class distributions from our medline-derived dataset are plotted in Figure~\ref{fig:mesh_dist}. As shown on the histogram, lots of classes are annotated in only one document, which makes them difficult to learn.

Another obstacle (independent from the pandemic) is the scale of the MeSH thesaurus, as there are thousand of MeSH descriptors. State-of-the-art on MeSH classification thus uses IR techniques \citep{article2017mesh}, or focuses on only single MeSH descriptors \citep{10.1145/3411408.3411414}.

In this work we rely on BioBERT \cite{Lee_2019} to extract representations from paper abstracts and classify them. Such model is pretrained with masked language modeling objectives on data from the biomedical domain, and we assume that BioBERT encodes some semantic knowledge related to the biomedical domain. However, it has been shown that this pretraining might not be optimal for tasks such as NER or NLI~\cite{jin2019probing}.

We formulate the zero-shot task as an ``open input" problem where we take both the class and the text as an input, and output a matching score between the two. The motivation behind this formulation is that the model would learn to use the semantics of the class labels, and thus will be able to extend the semantic knowledge encoded by pretrained model (eg. BioBERT) to new classes. Therefore, our assumption is that those models host and can make use of a good representation of the semantics underlying the MeSH hierarchy, including unobserved terms.

\begin{figure}
	\includegraphics[width=0.9\columnwidth]{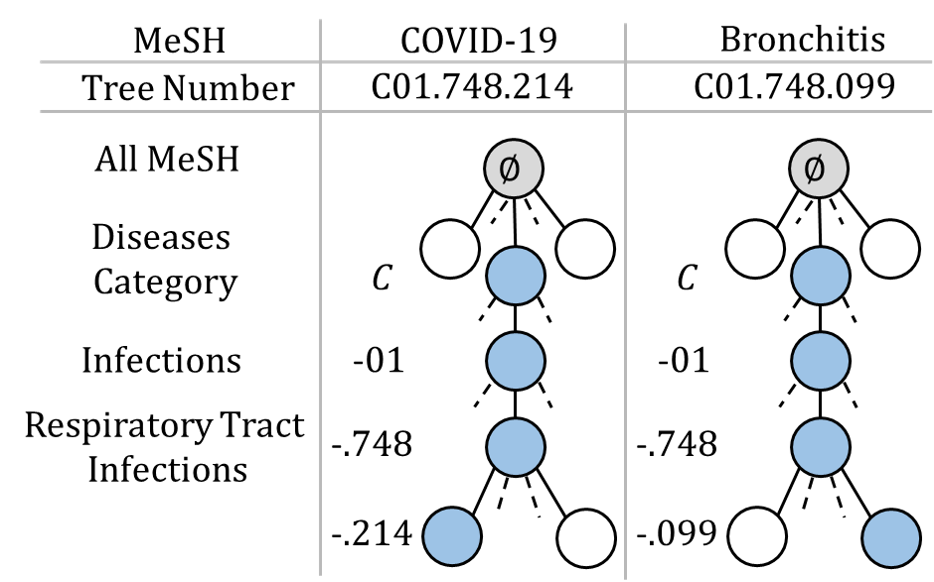}
	\caption{Example of the MeSH hierarchy. \textit{COVID-19} and \textit{Bronchitis} have three common ancestors (\textit{Diseases category}, \textit{Infections} and \textit{Respiratory Tract infections}) and then split in distinct MeSH descriptors. The hierarchy is all included in the Tree Numbers of the MeSH terms.}
	\label{fig:mesh_ex}
\end{figure}

In order to improve the semantic representations of pretrained model, we propose a multi-task training framework, where an additional decoder block predicts the position in the hierarchy of the input MeSH term.  MeSH descriptors have a position in the hierarchy that is defined by their Tree Numbers (see Figure \ref{fig:mesh_ex}), so the goal of this secondary module would be to generate those Tree Numbers during training. Model learnt with this additional task should better encode MeSH hierarchical structure and hopefully improve zero-shot and few-shot capacities of thus learnt representations. Enforcing that semantic knowledge is embedded in the model also guarantees a degree of explainability, an important feature in the medical domain.

The main findings of the work are: (a) Our multi-task framework improves precision on some datasets, and thus the F1-score, but it is not systematic. (b) Probing tasks show that performance increases are directly linked to a better knowledge of the MeSH hierarchy. Still, including hierarchical information on a large scale dataset (Medline) is difficult, especially in few-shot and zero-shot settings.

\section{Related work}

\paragraph{Zero-shot classification.} There is a large literature on zero-shot learning, which consists in classifying samples with labels not seen in training~\cite{wang2019survey,chen2021knowledge}. Pre-trained models such as BERT \cite{devlin2019bert} or BioBERT \citep{ Lee_2019} are central to zero-shot learning in NLP. These models benefit from very large datasets they can be trained on in a self-supervised way. Such pretraining allows them to learn rich semantic representation of the text, and perform knowledge transfer on other lower-resourced tasks. Those pre-trained models can be used in a zero-shot setting, by creating representations for the given document and each of the the different classes, and then computing similarity scores based on those representations. \citet{chalkidis2020empirical} proposed for example to compute the similarity score with an attention mechanism between classes and documents representations. \citet{rios-kavuluru-2018-shot} proposed in addition to the attention mechanism to include hierarchical knowledge using GCNN, but they do not handle the case where the hierarchy is only available during training. \citet{wohlwend2019metric} also worked on the representation space using prototypical network and hyperbolic distances and they showed that there was still possible improvements in metric learning. 

\paragraph{Fine-grained biomedical classification.} BioASQ challenge is one of the reference on fine-grained classification of biomedical articles; however the challenge does not focus on zero-shot adaptation, which is the scenario we consider in this work.  \citet{10.1145/3411408.3411414} have tried to perform zero-shot classification across MeSH descriptors, but their testing settings considered only a small number of MeSH descriptors. In our work we try to perform a larger scale evaluation in context of the pandemic. Finally, \cite{inproceedings} proposed an architecture for hierarchical classification tasks that is able to learn the hierarchy by generating the sequence from the hierarchy tree (using an encoder/decoder architecture). Our work considers similar architecture in a zero-shot scenario.

\paragraph{Probing.} Probing models \citep{conneau2018cram, tenney2019learn} are lightweight classifiers plugged on top of pretrained representations. They allow to assess the amount of ``knowledge" encoded in the pretrained representations. \citet{alghanmi2021probing, jin2019probing} introduced frameworks in the biomedical domain for disease knowledge evaluation focusing on relation types (\textit{Symptom-Disease}, \textit{Test-Disease}, etc.), while we are willing to assess how well a hierarchical structure is encoded in representations.
We rely on the structural probing framework~\citep{hewitt-manning-2019-structural} that we compare against the hierarchical structure encoded by MeSH thesaurus.
 
\section{Proposed approach}

\begin{figure*}
    \centering
	\includegraphics[width=0.8\textwidth]{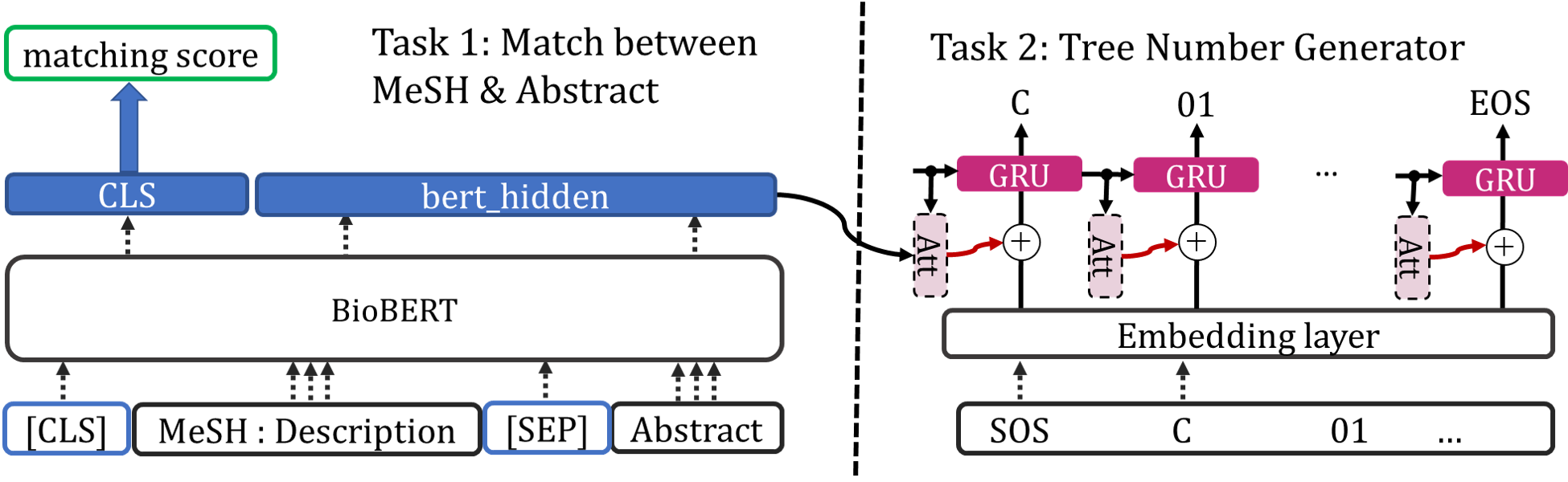}
	\caption{Architecture \textbf{BioBERT MTL}. The encoder (blue) creates a representation of the MeSH, then the decoder (red) generates the Tree Number from this representation. GRU cells are used in the decoder as well as an attention mechanism to better handle long sequences. The binary output (green) is the matching score.}
	\label{fig:decoder}
\end{figure*}

First, we will explain the architectures we explored to address the zero-shot classification problem, and more precisely the multi-task learning framework. Then, we will focus on the design of the probing tasks, that we used to analyse to what extent the hierarchical knowledge was encoded by the different representations. 

\subsection{Zero-shot Architecture}
\label{sec:models}

\paragraph{BioBERT Single-Task Learning (STL).} The first model is a BioBERT encoder, followed by a dense layer on the {\tt [CLS]} token. Input of BioBERT is composed of the MeSH term, the MeSH description and a document abstract: {\tt [CLS] MeSH term : MeSH description [SEP] Abstract}, and output is a single neuron, that goes through a sigmoid activation function.  

As an example, input for the MeSH term \textit{Infections} is {\tt [CLS] Infections : Invasion of the host organism by microorganisms or their toxins or by parasites that can cause pathological conditions or diseases. [SEP] Abstract}.

\paragraph{BioBERT Multi-Tasks Learning (MTL).} The second architecture is similar to the BioBERT STL, but in addition to the binary classification task it learns simultaneously an additional task of MeSH term hierarchical position generation. The motivation behind this additional task is that the learnt representations would better encode hierarchy of MeSH terms and hopefully better deal with zero-shot classification or fine-grained classification problems.

Figure \ref{fig:decoder} describes the architecture of this model. It uses an additional decoder block, as a secondary task to predict the tree number of the given input.
The generation step $j+1$ is defined as (without considering the batch size):
\begin{align}
& att_j = bert\_h \times h_j \\
& \widehat{att_j} = softmax(att_j) \\ 
& attn\_applied_j = \widehat{att_j}^T \times bert\_h \\
& input_j = embed_j + attn\_applied_j \\
& h_{j+1}, out_{j+1} = GRU(h_j, input_j)
\end{align}
where $bert\_h$ is the output of BERT, of shape (512, 768), $h_j$ the hidden state of the GRU cell, of shape (768,), and $embed_j$ the embedding of the current word, also of shape (768,). On line (4), the $+$ operator corresponds to a sum of the two vectors (both of shape (768,)) in each of their dimensions. For the word generation, $out_{j+1}$ is then passed through a dense layer and a logsoftmax function.

Note that $bert\_h$ is formed from the output tokens of BioBERT corresponding to the MeSH description (by applying the MASK to all other tokens). We also apply ``teacher forcing", to reduce error accumulation.

The original problem is thus transformed in a multi-tasks problem, where the two losses (binary cross entropy and negative log likelihood losses) are then jointly learned:
\begin{align}
loss_{tot} &= \frac{1}{2\sigma_1^2} loss_1 + \frac{1}{2\sigma_2^2} loss_2 + \log(\sigma_1 \sigma_2)
\end{align}
where both $\sigma_1$ and $\sigma_2$ are learnable parameters included in the models parameters \citep{8848395, kendall2018multitask}, to allow the model to balance between the binary and tree number generation losses. The last regularization term is only here to prevent the model to learn the naive solution of just increasing $\sigma_1$ and $\sigma_2$ to reduce the loss.

The output vocabulary of the decoder is composed of the tree numbers tokens: {\tt A-Z} letters, {\tt 00-99} digits, and {\tt 000-999 } digits. All together, the vocabulary size is around 1100, on which we apply an embedding layer to transform discrete tree numbers to a continuous embedding space. As this vocabulary is completely new, embedding is learnt from scratch using back-propagation. Note that for MeSH descriptors that have multiple tree numbers, we just duplicate the inputs to learn the multiple positions in the hierarchy.

In both architectures, the full set of parameters is updated during training. Thus, each model provides new representations of the input text and labels, that are further evaluated through zero-shot classification or through probing.

\subsection{Hierarchical Probing Task}
\label{sec:probe}

To better understand the capacity of pretrained representations to encode hierarchical relations of biomedical terms, we considered two probing tasks, adapted from \citep{hewitt-manning-2019-structural}. The objective is to test whether the representations learned by the model are linearly separable with regards to the probing tasks.

We define the probing task from a main model, taking as input a MeSH descriptor $\textbf{m}_i$ and returning its internal representation $\textbf{h}_i$. We then recall that it is possible to define a scalar product $\textbf{h}^T A \textbf{h}$ from any positive semi definite symmetric matrix $A \in \mathbf{S}_+^{m\times m}$, and more generally from any matrix $B \in \mathbf{R}^{k\times m}$ by taking $A=B^T B$. Using the metric distance corresponding to this scalar product we then define a distance from any matrix :
$$
d_B(\textbf{h}_i, \textbf{h}_j)=(B(\textbf{h}_i-\textbf{h}_j))^T(B(\textbf{h}_i-\textbf{h}_j))
$$
with $\textbf{h}_i$ and $\textbf{h}_j$ the representations of two MeSH descriptors (more details on representations in section \ref{sec:probe_results}). Our model has as parameter the matrix $B$, which is trained to reconstruct a gold distance from one MeSH term to another. More specifically, the task aims to approximate by gradient descent :
$$
\underset{B}{min} \sum_{i,j} |d_T(\textbf{h}_i, \textbf{h}_j) - d_B(\textbf{h}_i, \textbf{h}_j)^2|
$$
where $d_B$ is the predicted distance and $d_T$ the gold distance. We also note that, as in the original paper, we add a squared term on the predicted distance. Concerning the dimensions of $A$ and $B$, $m$ is the dimension of the representation space (same than \textbf{h}), and $k$ is the dimension of the linear transformation which we will take equal to 512. We did not further experiment on the dimension of the linear transformation (see the original paper by \cite{hewitt-manning-2019-structural} for discussion on both the squared distance and the linear transformation dimension).

\paragraph{Gold distance.} The only difference with the original paper is the definition of the gold distance. We have evaluated two probes: 
\begin{itemize}
    \item \textbf{Shortest-Path Probe}: given two MeSH descriptors, we ask the model to predict the distance between the two MeSH terms, as the length of the shortest path in the graph defining the MeSH hierarchy;
    \item \textbf{Common-Ancestors Probe}: model predicts whether two MeSH descriptors have k common ancestors. For this second task we thus define multiples binary probe models that predict if the two MeSH terms have at least k common ancestors or not (for k between 1 and 3). In this particular case of a binary probe task, we thus add a sigmoid function on the predicted distance (where the sigmoid function is not centered on zero, but on a positive constant, as distances are always positive)\footnote{We have also tried to cast the probe as a regression directly predicting the number of common ancestors given two MeSH descriptors, but the regressor was unable to train from the representations, hence the use of binary tasks.}.
\end{itemize}

\section{Experimental settings}

In this section we present the datasets we used to train the models and evaluate the corresponding representations. We also explain how we construct a zero-shot dataset out of the Medline dataset with MeSH annotations. 

\subsection{Datasets}
\label{sec:dataset}

\paragraph{Medline/MeSH.} Medline is the US National Library of Medical/Biomedical dataset\footnote{\url{https://www.ncbi.nlm.nih.gov/mesh/}}, containing millions of biomedical scientific documents, and built around the MeSH thesaurus (Medical Subject Heading). This thesaurus contains about 30,000 MeSH descriptors, updated every year, and used for semantic indexing of the documents. These MeSH terms also define a hierarchy: the first level separates the MeSH terms into 16 main branches, then each MeSH term is the child of another more general MeSH term, and this over up to a depth of fifteen. The sequence of nodes traversed to reach a MeSH term from the root is called Tree Number.

An example of the hierarchy is shown in Figure \ref{fig:mesh_ex} with the two MeSH \textit{COVID-19} and \textit{Bronchitis}.  For example here, \textit{Covid-19} has the Tree Number {\tt C01.748.214} ({\tt C} being the main branch \textit{Disease}, then we have 3 sub-levels: {\tt C01} for \textit{Infections}, {\tt C01.748} for \textit{Respiratory Tract Infections}, then {\tt C01.748.214} for \textit{Covid-19}).

The majority of MeSH descriptors from the hierarchy have multiple Tree Numbers, so the hierarchy follows a directed acyclic graph structure. Also, the annotation of scientific documents with ancestors of a MeSH is not always explicit. For example, a document can be indexed with term \textit{Covid-19}, but not necessarily with terms  \textit{Infections} or \textit{Respiratory Tract Infections}. 

There are on average 13 annotated MeSH descriptors per  document, where 2 or 3 will be annotated as \textit{major MeSH} to indicate that the document deals more specifically with these topics.  In our work, we use the whole set of major and non-major MeSH descriptors. In addition to the MeSH annotation and hierarchy, the Medline database provides a description for each MeSH term, used by our models as specified in section \ref{sec:models}.

\paragraph{LitCovid.} LitCovid is a subset of the Medline database \citep{RN12503, chen2020litcovid}, where extraction is done via PubMed (search engine of Medline), using the keywords: ``coronavirus", ``ncov", ``cov", ``2019-nCoV", ``COVID-19" and ``SARS-CoV-2". Using this subset of articles allow us to work more specifically on COVID-19 related articles, with also a subset of 9,000 COVID-19 related MeSH descriptors (instead of the full set of MeSH descriptors). The LitCovid dataset also contains its own categorization, composed of only 8 classes: Case report, Diagnosis, Forecasting, General, Mechanism, Prevention, Transmission and Treatment, with as for the MeSH a short description for each of them.

All our experiments are made on the LitCovid dataset, with 27,321 articles (Train-Val-Test split: 19,125 / 2,732 / 5,464) that have both LitCovid and MeSH annotations (several of the 8 classes from LitCovid + avg 13.5 MeSH/article out of around 9,000 COVID-19 related MeSH descriptors from Medline). 

Training and evaluation relies on MeSH annotations (semantically richer), with results that reflects both few-shot for low frequency terms, and zero-shot results for 747 held-out MeSH descriptors. In a second step we also evaluate on LitCovid categories to test a transfer learning scenario where we change the categorization at test time.

\subsection{Evaluation}

We present in this section the adaptation of annotations for the ``open input" architectures. The objective is to create pairs ($\{label, document\}, boolean$) from the original annotation. 

\paragraph{Zero-shot dataset creation.} Inputs in zero-shot are different due to new class appearances: a document $d_1$ being associated with two labels ($l_1$ and $l_2$) will thus be transformed into two inputs ($\{l_1,d_1\}$, positive) and ($\{l_2, d_1\}$, positive). When a new label $l_{new}$ appears, it is enough to create the input ($\{l_{new}, d_1\}$) to predict whether this label $l_{new}$ is positive or not for the document $d_1$.

\begin{table*}
    \centering
    \begin{tabular}{l c c c | c c c}
         & F1-score (std) & Precision & Recall & F1-score (std) & Precision & Recall \\
        \hline
        \hline
         & \multicolumn{3}{c}{Balanced} & \multicolumn{3}{c}{ZSC Balanced} \\
         \hline
         \hline
        BioBERT STL & \underline{\textbf{0.894}} (0.001) & \underline{\textbf{0.875}} &\underline{ \textbf{0.913}} & \underline{\textbf{0.760}} (0.004) & 0.849 & \underline{\textbf{0.688}} \\ 
        BioBERT MTL &  0.873 (0.003) & 0.868 & 0.878 & 0.754 (0.006) & \underline{\textbf{0.856}} & 0.674 \\
        \hline
        \hline
        & \multicolumn{3}{c}{Siblings} & \multicolumn{3}{c}{ZSC Siblings} \\
        \hline
        \hline
        BioBERT STL & 0.390 (0.002) & 0.300 & 0.562 & 0.281 (0.005) & 0.558 & \underline{\textbf{0.470}} \\ 
        BioBERT MTL & \underline{\textbf{0.402}} (0.005) & \textbf{\underline{0.312}} & \underline{\textbf{0.570}} &\underline{\textbf{0.285}} (0.005) & \underline{\textbf{0.594}} & 0.455 \\
        \hline
    \end{tabular}   
    \caption{Results on Medline/MeSH with different evaluation configurations. Training has been done on the {\tt balanced} dataset, while here we test on both {\tt balanced} and {\tt siblings} dataset. In {\tt zsc balanced}, all MeSH descriptors are zero-shot, while in {\tt zsc siblings} it is a generalized zero-shot setting (mixture of both zero-shot and non zero-shot MeSH descriptors).}
    \label{tab:prec/rec}
\end{table*}

To make the task meaningful, we also add negatives to both train and test datasets. However, in the case of MeSH classification, using all the negatives is not possible, for two reasons: 
\begin{enumerate*}[label=(\roman*)]
    \item scalability problem: there are more than 9,000 labels and 27,321 documents, that results in hundreds of millions of combinations.
    \item data balancing problem: 9,000 negatives for 13 positives on avg.
\end{enumerate*}
We therefore use following configurations:
\begin{itemize}
    \item {\tt Balanced}: one random negative pair is added for each positive pair, to ensure a balanced distribution. So, given a document, we would always have the same number of positive and negative pairs. The negatives are sampled from all the possible negatives, based on the MeSH terms distributions to ensure that, given a MeSH term, we would also have the same number of positive and negative pairs.
    \item {\tt Siblings}: This configuration is only used in evaluation and aims at better disentangle errors due to ``incompleteness of MeSH annotations" from real indexing errors. In this configuration, siblings according to MeSH hierarchy of the positive pairs are added with negative labels. We also add all the ancestors of the annotated MeSH terms as positive labels to overcome annotation incompleteness problem stated above. Adding the ancestors increases the size of the dataset by an important factor, so this is why this configuration is used for evaluation only.
\end{itemize}

The choice of the negatives is crucial in metric learning, and there have been lots of efforts given on developing techniques to find ``hard negatives". In our case, the {\tt Siblings} configuration creates by its nature negatives that are difficult to distinguish from actual positives. Also we made the choice to use a binary classification layer, but losses like hinge loss or triplet loss could have been interesting in this particular case.

For LitCovid we consider all the \{document, label\} pairs since we do not have scaling problem (only 8 labels).

\subsection{Training parameters}

The losses (binary cross entropy for the binary task and negative log-likelihood for the hierarchy generation task) are optimised using the AdamW and Adam algorithm with a learning rate of 2E-5 and 5E-4 respectively. Training is done over 4 epochs, with a save every 0.25 epochs. Best model is selected based on the validation loss. We used a batch size of 16, and performed 3 runs for each model (see standard deviation in the section \ref{sec:results}).

The BioBERT pretrained model we used was {\tt monologg/biobert\_v1.1\_pubmed} from Hugging Face. This model accepts an input sequence of up to 512 tokens, therefore extra tokens were truncated.

Concerning probing tasks, each MeSH-to-MeSH pair requires a gold distance (see section \ref{sec:probe}). For the \textbf{Shortest-Path Probe} they are computed using the Floyd-Warshall algorithm, while for the \textbf{Common-Ancestors Probe}, they were deducted from Tree Numbers. Optimizer of the probe task is AdamW, with a 2.5E-5 learning rate. We also only focused on the ``N", ``E", ``C", ``D" and ``G" branches of the MeSH hiearchy, corresponding resp. to ``Health Care Category", ``Analytical, Diagnostic and Therapeutic Techniques and Equipment Category", ``Diseases Category", ``Chemicals and Drugs Category" and ``Phenomena and Processes Category" (they are the most representative of the dataset). From all possible MeSH-to-MeSH pairs, we have randomly selected 10\% of them to reduce computation time. Validation and evaluation are performed on 30\% of the MeSH descriptors, that we held out from probe training.

\section{Results and discussion}
\label{sec:results}
We present in this section the results in zero-shot and few-shot on both MeSH descriptors and LitCovid labels from our two models BioBERT STL and MTL. We then discuss the results of the probing tasks, the architectures, the quality of annotations and also how we approached the problem of large scale zero-shot classification.

\subsection{Zero/Few-shot classification}

\paragraph{Medline/MeSH.} The models have been trained on the {\tt balanced} configuration, and then tested on {\tt balanced} and {\tt siblings}. Table \ref{tab:prec/rec} compares results on those different test set configurations both in non zero-shot and zero-shot settings. Note, that as highlighted in section \ref{sec:dataset}, the {\tt siblings} configuration is more difficult than the {\tt balanced} one, which explains  a high gap between the F1-score from the two configurations. This is mainly due to a lower precision as the model tends to wrongly predict the siblings from the positive MeSH terms as positives examples (while we consider them as negatives in {\tt siblings} settings). On the {\tt balanced} test set, BioBERT STL has a better F1-score both in zero-shot and non zero-shot settings, while on the {\tt siblings} one, the BioBERT MTL model performs better. This difference is due to the high precision of BioBERT MTL model in both settings. More precisely, the BioBERT MTL model seems to be better on difficult pairs, like in the {\tt  siblings} settings where you may have very close negative pairs (for example \textit{Breast Cyst} positive and \textit{Bronchogenic Cyst} negative).

\begin{figure}
    \centering
	\includegraphics[width=0.9\columnwidth]{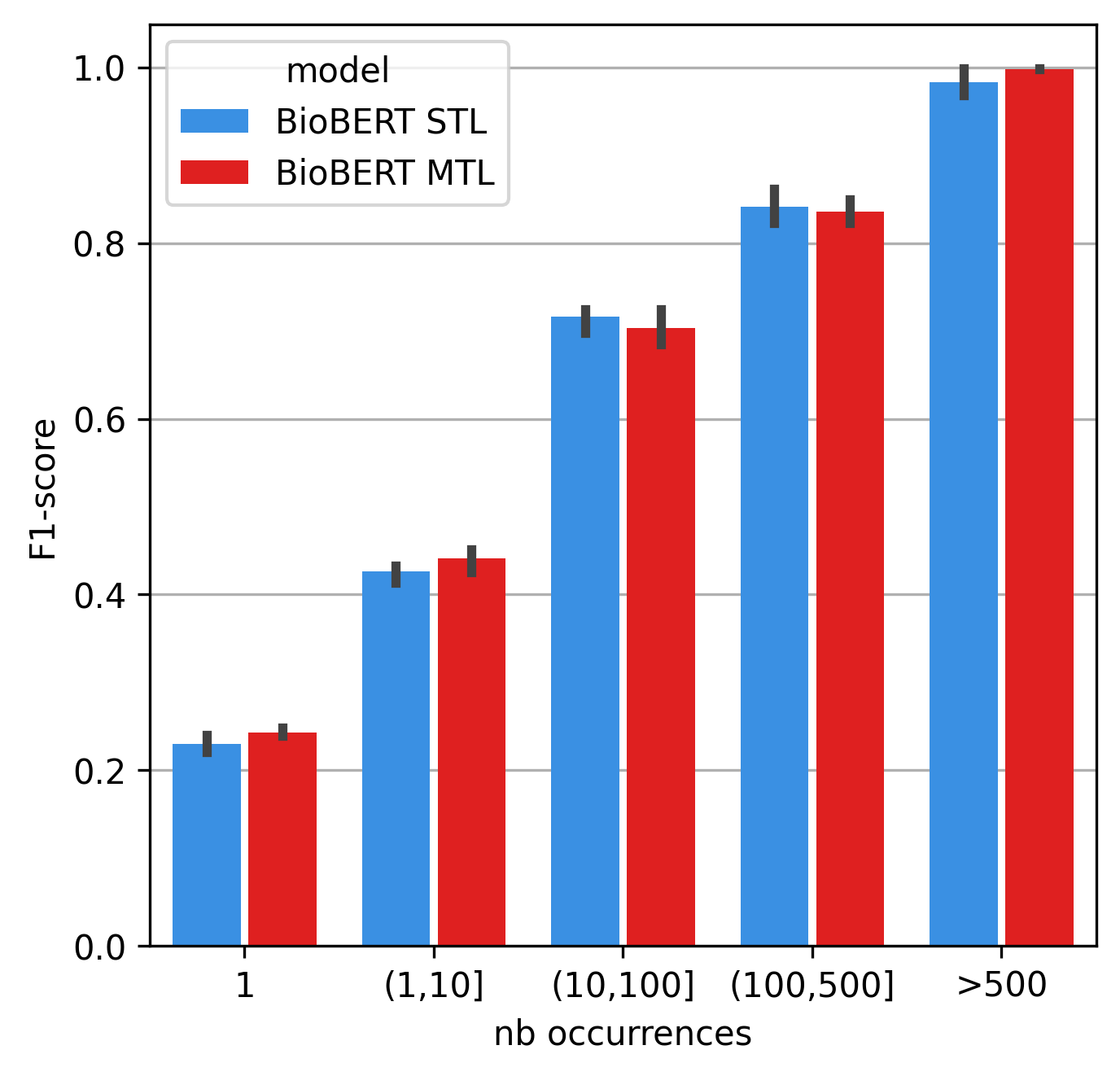}
	\caption{F1-score depending on the number of occurrences of the MeSH term in the train set. Evaluation dataset is {\tt siblings}, training dataset is {\tt balanced}.}
	\label{fig:res_freq}
\end{figure}

 Figure \ref{fig:res_freq} plots the F1-score with respect to the number of occurrences of the MeSH descriptors in the train set thus allowing to evaluate few-shot learning quality. First we note a clear increase of F1-score with the number of occurrences of the MeSH terms in the dataset (up to 0.7) on both models. This indicates, as one would expect, that the models are really struggling with difficult pairs that contains rare MeSH descriptors. In comparison with Table \ref{tab:prec/rec}, the F1-score in zero-shot on the balanced pairs is of $0.76$, while the F1-score for the rare MeSH terms is much lower. We also note, that the BioBERT MTL model allows to  slightly improve performance in low resources settings (for the terms occurring less then 10 times in the training data: $1$ and $\left(1, 10\right]$ bins in the figure).

\begin{figure}
	\includegraphics[width=\columnwidth]{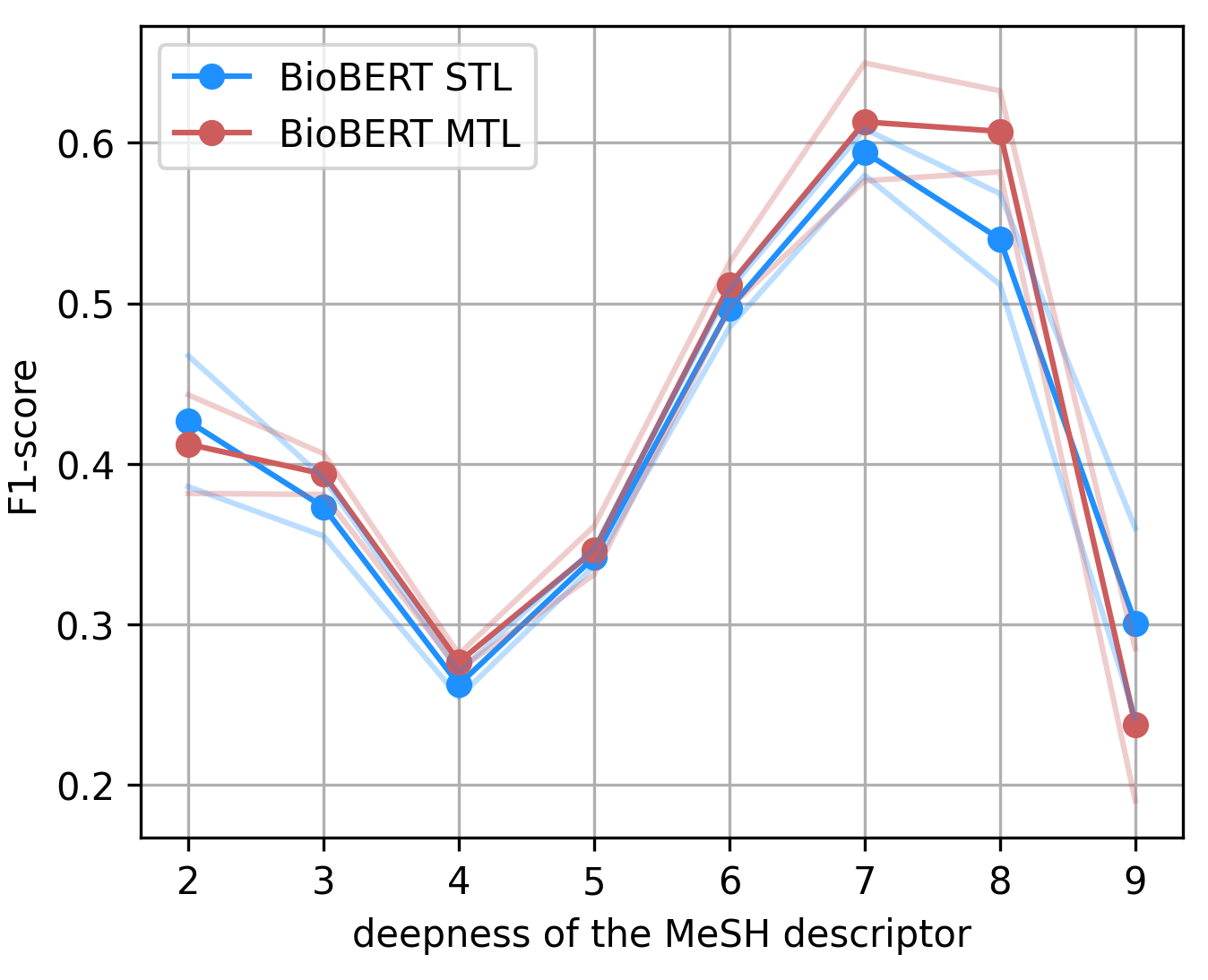}
	\caption{Comparison of F1-score depending on the deepness of the MeSH descriptors on both BioBERT STL and MTL models. Test set is {\tt siblings}, and depth is computed from the length of the MeSH descriptor Tree Numbers (average depth when MeSH terms have multiple Tree Numbers).}
	\label{fig:deep}
\end{figure}

Concerning the MeSH descriptors themselves, Figure \ref{fig:deep} shows F1-scores with respect to the deepness of the MeSH descriptors both for BioBERT STL and MTL models. As shown in the figure, F1-score tends to decrease for more general terms (first 4 levels of hierarchy), but increases for more specific terms (after 4th level of hierarchy). We believe this could be due to the incompleteness of annotations used during training of the MeSH descriptors. Recall that training is performed in the {\tt balanced} setting, and therefore ancestor MeSH descriptors are not always explicitly annotated, which could result in low performance on high hierarchy levels. This graph is also difficult to interpret since some branches from the MeSH hierarchy are deeper than others, therefore ``specificity" of term with respect to its absolute depth may be different; the only information on depth is that deeper MeSH terms are in general more specific one.

\paragraph{LitCovid.} Table \ref{tab:litcovid} reports results on LitCovid dataset in zero-shot setting for the representations obtained with STL and MTL models. On LitCovid, the STL model is better. We believe this  may be due to LitCovid categories being very general in comparison to the MeSH descriptors, therefore the MTL model could not take advantage of its better precision on more specialised pairs. In addition, as previously, this could be due to the incompleteness of MeSH annotations, where only most specific MeSH terms are present in training data, while LitCovid relies on more generic labels. 

We report two simple baselines for LitCovid dataset in Table \ref{tab:litcovid}: \begin{itemize}
    \item \textit{Baseline IsIn} where an abstract is associated with a label that appears in the abstract itself (both lower-cased);
    \item \textit{Baseline Cos Sim}, where we take the {\tt CLS} token representations of all labels (through a vanilla BioBERT), same for all abstracts, and then compute the cosine similarity between each pair, with a threshold defined on the validation set.
\end{itemize}
We note that both BioBERT-based models perform significantly better compared to those naive baselines, which indicates that the models are able to exploit the semantics of the label to some extent, and goes beyond simple label lookup in the abstract. 
We also see that the increase of F1-score is mainly due to a better recall which implies better coverage of our models.

\subsection{Probing hierarchy knowledge}
\label{sec:probe_results}

Finally, Table \ref{tab:probe} reports the results of probing the learnt representations. Results for the \textbf{Shortest-Path Probe} are based on the mean error on the predicted distance with respect to the gold distance (shortest path length between two MeSH terms), while for the \textbf{Common-Ancestors Probe} we report F1-score on binary tasks for each $k$ (evaluating whether the two MeSH have at least $k$ common ancestors).

\begin{table}
    \centering
    \begin{tabular}{l | c c c}
        \hline
        \hline
        ZSC LitCovid & F1-score & Precision & Recall \\
        \hline
        \hline
        Baseline IsIn & 0.329 & 0.520 & 0.241 \\
        Baseline Cos Sim & 0.308 & 0.228 & 0.471 \\
        \hline
        BioBERT STL & \underline{\textbf{0.512}} & \underline{\textbf{0.444}} & \underline{\textbf{0.604}} \\ 
        BioBERT MTL & 0.465 & 0.401 & 0.553 \\
        \hline
    \end{tabular}   
    \caption{Results in zero-shot on LitCovid, where the model has been trained on the MeSH descriptors ({\tt balanced}). F1-score is the best from three runs.}
    \label{tab:litcovid}
\end{table}

\paragraph{MeSH representations.} We use {\tt CLS} token of respective models as representation of different MeSH in our experiments. We have compared this representation to both average pooling over the MeSH descriptors tokens and max pooling in our preliminary experiments, and observed that {\tt CLS} token was leading to the best performance on the probe tasks.

As a baseline, we give in table \ref{tab:probe} two other representations: \textit{BioBERT vanilla} and \textit{Random}. In \textit{BioBERT vanilla}, representations are an average pooling of the MeSH output tokens provided by the BioBERT pretrained model without any finetuning (avg pooling was in this only case better than the {\tt CLS} token\footnote{possibly because in \textit{BioBERT vanilla} model {\tt CLS} token representations has not been finetuned for any task as opposed to our learnt models.}), and for \textit{Random}, MeSH representations are random representations sampled from a normal distribution.

\paragraph{Comparison STL/MTL.} Table \ref{tab:probe} indicates that both STL and MTL model encode hierarchical structure of MeSH terms better than random baseline, but also better that BioBERT vanilla baseline. More specifically, the \textbf{Common-Ancestors Probe} implies that between two MeSH descriptors from the same categories, we have encoded a common base, and there is a projection where MeSH descriptors from the same categories are closer to each others. Concerning the \textbf{Shortest-Path Probe}, results shows that there is also a projection where distances (as shortest paths in the hierarchy graph) are respected. From the results, we also see that the additional task, MTL model with the decoder block, is able to encode even more hierarchical information in the {\tt CLS} token, and may be a hit to the better precision in zero-shot and few-shot results. Also, it is interesting to see that in the \textit{BioBERT vanilla} model, there is already some good knowledge about this hierarchical structure, which makes sense as this hierarchy is constructed on the semantics of the biomedical terms.

\subsection{Limitations and possible future directions}

\paragraph{Multi-Tasks-Learning.} When dealing with MTL, the main difficulty comes from convergence speed of the different losses. In our framework, the main loss (classification loss) converges faster than the secondary loss (decoder loss), so we are not able to take full advantage of the decoder architecture. In a perfect scenario, the main task should be the harder one, but here it was not the case, so we were forced to stop training earlier even when using different coefficients and learning rates for the two losses. Another possible future direction to explore is to start training the decoder block before training the classification layer. 

\begin{table}
    \centering
    \begin{tabular}{l | c c c }
        \hline
        \hline
        \textbf{Shortest-Path Probe} & \multicolumn{3}{c}{Distance Error (std)} \\
        \hline
        \hline
        BioBERT vanilla &\multicolumn{3}{c}{ 1.494 (-)} \\
        Random & \multicolumn{3}{c}{2.597 (-)} \\
        \hline
        BioBERT STL & \multicolumn{3}{c}{1.462 (0.053)} \\ 
        BioBERT MTL & \multicolumn{3}{c}{\underline{\textbf{1.323}} (0.013)} \\
        \hline
        \hline
        \textbf{Common-Ancestors Probe} & F1 (k=1) & k=2 & k=3 \\
        \hline
        \hline
        BioBERT vanilla & 0.855 & 0.521 & 0.538 \\
        \hline
        BioBERT STL & 0.864 &0.546 & 0.541 \\ 
        BioBERT MTL & \underline{\textbf{0.933}} & \underline{\textbf{0.659}} & \underline{\textbf{0.576}} \\
        \hline
    \end{tabular}   
    \caption{Results of the probe tasks. For comparison of the \textbf{shortest-path} task, the avg MeSH to MeSH distance is 10.033, with a std of 3.016. For the \textbf{common-ancestors} task, k is the number of common ancestors.}
    \label{tab:probe}
\end{table}

\paragraph{Annotations.} When dealing with transfer learning across different datasets, the question of the quality of the annotation needs to be taken into account. Different annotation systems (even when documents are manually annotated) may have labels that have different coverage, and overlapping, which adds some bias in results. As an example, when we train our model on the Medline annotations and then test in zero-shot on LitCovid labels, results are difficult to interpret, because the scale and the coverage is completely different. \citep{MINARROGIMENEZ201937} have studied the semantic interoperability of different biomedical annotation tools across multiple countries and databases, and they show that this was a real issue, that needs to be considered when dealing with such terminologies.

\paragraph{Large scale Zero-shot.} ``Open input" architectures are not adapted to very large scale zero-shot problems, as we need to create too many pairs for a given document. Our approach was to work on a balanced subset of the possible pairs or a coherent one (resp. balanced and siblings configurations) for training and evaluation, however, this technique does not adapt to real world applications. 

An interesting future direction could be combining our ``open input" architecture with the high-coverage retrieval-like step which would first pre-select a subset of possible MeSH terms, and therefore restrict the search space for the second ``open input" classification step. For example of the first step, a ColBERT model could compute representations of abstracts and classes independently instead of creating representations of pairs, and so reduce computational cost. Another possibility could be to use simple algorithms like BM25.

Other techniques in metric learning also exist, like triplet loss learning or hinge loss. Using triplets with ``hard negatives" may help to learn a better representation space.

\section{Conclusion}
In this work, we try to address the problem of zero-shot classification that we defined as an \textit{open-input} problem. We compare a simple BioBERT model with a multi-tasks learning architecture that  includes hierarchical semantic knowledge. In zero-shot and few-shot settings, the multi-tasks framework does not increase performances significantly. Still, we observe good results on precision and on structural probing tasks, which implies that the addition of the seq2seq task has some beneficial effect in the ability of trained models to capture semantics. In particular, the model is able to build a representation space where MeSH descriptors that have common ancestors are closer to each other, and where the overall hierarchical organisation of the MeSH is respected. It would be interesting to further investigate additional tasks to take even better advantage of hierarchical knowledge encoded in medical terminologies, and thus improve quality and robustness of models representations. 

\bibliography{aaai22}

\end{document}